\documentclass[letterpaper, 10 pt, conference]{ieeeconf}
\IEEEoverridecommandlockouts 
\usepackage[letterpaper, left=54pt, right=54pt, bottom=54pt, top=54pt]{geometry}

\usepackage{graphicx}
\usepackage{amsmath} 
\usepackage{amssymb}  
\usepackage{mathtools}
\usepackage{bbm}
\usepackage{dsfont}
\usepackage{multirow}
\usepackage[dvipsnames]{xcolor}
\usepackage{subfig}
\usepackage[noadjust]{cite}
\usepackage{enumerate}
\usepackage{booktabs}
\usepackage{hyperref}

\usepackage{algpseudocode}
\usepackage{algorithm}
\usepackage{caption}
\usepackage[free-standing-units=true]{siunitx}
\usepackage{cleveref}

\title{\LARGE \bf
Towards Safe Multi-Level \\
Human-Robot Interaction in Industrial Tasks}

\author{Authors}

\author{Zhe Huang, Ye-Ji Mun, Haonan Chen, Yiqing Xie, Yilong Niu, Xiang Li,\\ Ninghan Zhong, Haoyuan You, D. Livingston McPherson, and Katherine Driggs-Campbell
\thanks{Z. Huang, Y. Mun, H. Chen, Y. Xie, Y. Niu, X. Li, H. You, D.L. McPherson, and K. Driggs-Campbell are with the Department of  Electrical and Computer Engineering and N. Zhong is with the Department of Computer Science at the University of Illinois at Urbana-Champaign. Emails: \{zheh4, yejimun2, haonan2, yiqingx2, yilongn2, xiangl5, ninghan2, hy19, dlivm, krdc\}@illinois.edu}%
\thanks{We thank Foxconn Interconnect Technology for supporting the authors with their research. This work was supported in part by ZJU-UIUC Joint Research Center for Cyber-physical Manufacturing Networks (CyMaN), Project  No. DREMES 202003, funded by Zhejiang University.}%
}

\begin{document}

\maketitle
\thispagestyle{empty}
\pagestyle{empty}

\begin{abstract}
Multiple levels of safety measures are required by multiple interaction modes which collaborative robots need to perform industrial tasks with human co-workers. We develop three independent modules to account for safety in different types of human-robot interaction: vision-based safety monitoring pauses robot when human is present in a shared space; contact-based safety monitoring pauses robot when unexpected contact happens between human and robot; hierarchical intention tracking keeps robot in a safe distance from human when human and robot work independently, and switches robot to compliant mode when human intends to guide robot. We discuss the prospect of future research in development and integration of multi-level safety modules. We focus on how to provide safety guarantees for collaborative robot solutions with human behavior modeling.
\end{abstract}
\section{Introduction}\label{sec:Introduction}

Collaborative robots (cobots) are increasingly deployed for high-mix low-volume production, since their effective interaction with co-workers improves productivity, and their flexibility allows fast adaptation to different task settings. Multiple levels of human-robot interaction are needed to qualify cobots for a wide range of industrial tasks. We introduce three safety modules corresponding to various interaction patterns: vision-based safety monitoring, contact-based safety monitoring, and hierarchical intention tracking. We discuss future directions on safety module integration and robust human behavior modeling.

\section{Safe Human-Robot Interaction Modules}\label{sec:safe-hri-modules}

We develop safe human-robot interaction modules using a UR5e arm, a Robotiq Hand-E Gripper and RGBD Intel RealSense cameras. The UR5e arm has joint torque sensors and end-effector force-torque sensors. All modules are executed on the Robot Operating System (ROS).

\subsection{Vision-based Safety Monitoring}
A traditional robot work cell typically sets up a physical barrier, or presence sensing devices such as light curtains and pressure sensitive safety mats. A cobot may be allocated to traditional tasks in which the cobot can work independently but with high speed. Thus, we develop a presence sensing module Vision-based Safety Monitoring (VSM). We use one top-down view camera to provide RGB frames of the shared space in the work cell. Human skeleton keypoints are detected by OpenPose~\cite{cao2017realtime} at 30 frames per second. The motion range of the cobot is specified in image space. If human skeleton position is detected within the motion range, the cobot execution is paused. The cobot execution is not resumed until the human skeleton position is detected outside the motion range. The advantages of VSM are its affordability and accessibility, since it only requires a RGB camera and the setup only involves hand-eye calibration. 

\begin{figure*}[t]
    \centering
    \includegraphics[width=\linewidth]{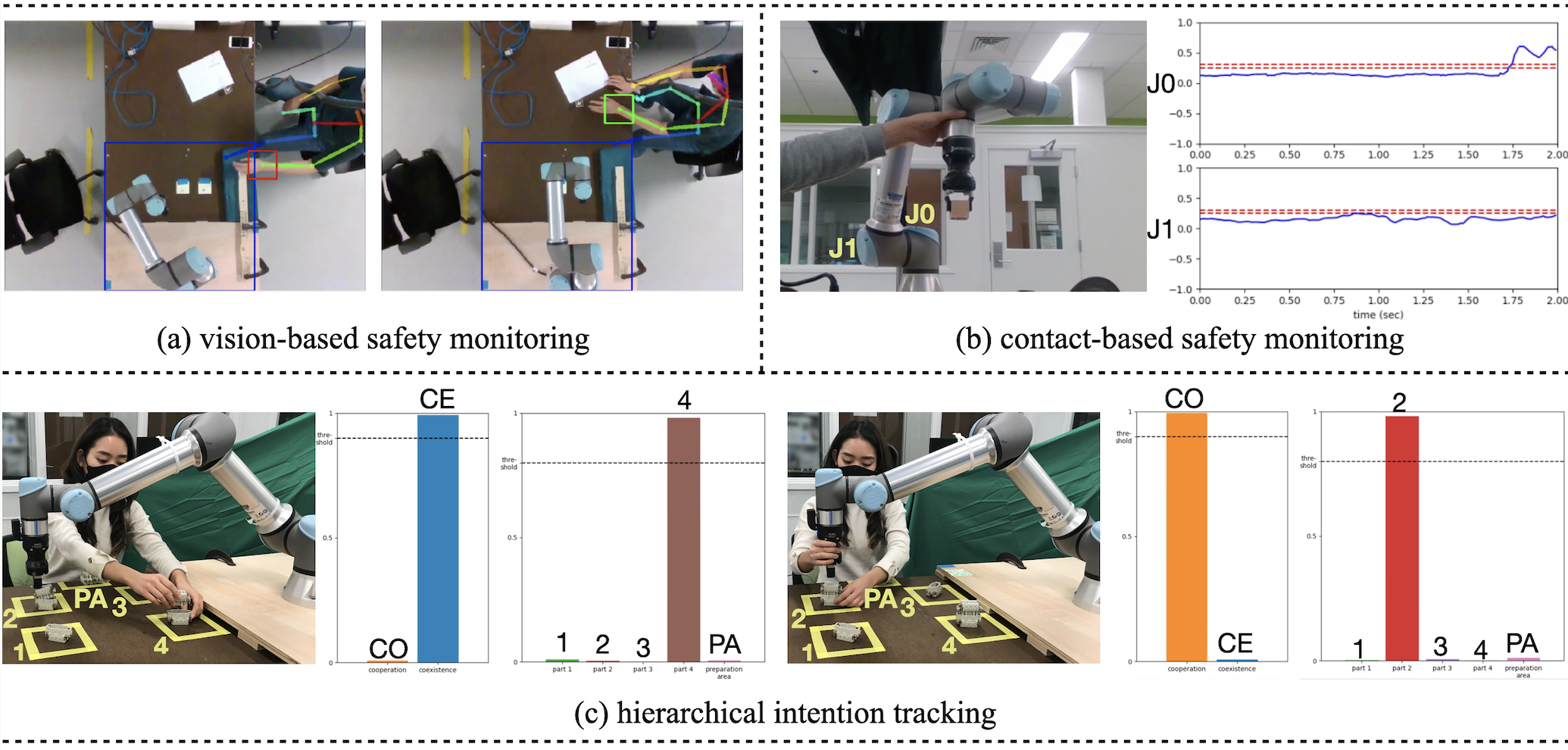}
    \caption{Demonstrations on safe human-robot interaction modules. (a) Vision-based safety monitoring: the cobot motion range is within blue rectangle. When human hand is inside the motion range, a red box is visualized and the cobot pauses the execution. When human hand is outside the motion range, a green box is visualized and the cobot resumes the execution. (b) Contact-based safety monitoring: we use torque measurements from joint 0 (J0) and joint 1 (J1) close to cobot base. The blue line is standard deviation value, and two dashed red lines are thresholds. (c) Hierarchical Intention Tracking: On the left, the human is doing her own job by taking a male part from preparation area (PA) to the 4th goal area and align it to a female part, and her interactive intention is tracked as coexistence (CE); On the right, the human is guiding the cobot to the 2nd goal area to fix the previous assembly failure, and her interactive intention is tracked as cooperation (CO).}
    \label{fig:safety-modules}
\end{figure*}

\subsection{Contact-based Safety Monitoring}
In collaboration tasks where the human co-worker needs to frequently visit area within the robot motion range, VSM sacrifices much efficiency by interrupting the cobot execution repeatedly. We develop Contact-based Safety Monitoring (CSM) to address safety in such tasks by pausing cobot execution upon unexpected contact detection. We use torque measurements on joints close to the cobot base for contact detection. The base joints are sufficient to capture contact on any part of the cobot, and torque measurements on base joints are more sensitive than on end joints. We calculate the standard deviation of the torque measurement over a 0.2 second time window at each joint, and set two thresholds. The contact is detected when the standard deviation of any joint torque exceeds the higher threshold, and the cobot execution is paused. The cobot execution is not resumed until the standard deviations of all joint torques keep lower than the lower threshold for more than 5 seconds. 

\subsection{Hierarchical Intention Tracking}
While humans are guaranteed by VSM and CSM to be safe to work at the same space with the cobot, productivity improvement is limited because neither VSM nor CSM allows human-centered interaction. The cobot always executes pre-defined trajectories, while the human co-worker needs to coordinate with the cobot's actions. To support more complicated and more free-form tasks, the human should take the lead of the human-robot team. Modeling human behavior is crucial for exploiting the potential of cobot, and it is essential to develop corresponding safety modules concurrently. We introduce Hierarchical Intention Tracking~(HIT), which performs human intention estimation on both interaction mode level and task level. The cobot uses the estimated human intention to switch execution between coexistence and cooperation interaction modes. During coexistence mode, the cobot and the human perform different subtasks separately. During cooperation mode, the cobot is guided by the human to recover task failures. For coexistence mode, artificial potential field generates robot motion repelled from human skeleton positions tracked by OpenPose, to keep a certain safe distance between the human and the robot. For cooperation mode, we implement admittance control to generate safe robot motion compliant to human guidance. More details are available in~\cite{huang2023hierarchical}.

\section{Future Work}\label{sec:future-work}
While VSM and CSM are ready to be directly applied in industry, collaborative robot solutions involving human behavior modeling such as HIT are still under active development. An inevitable concern is that human behavior modeling may yield inaccurate results due to noisy human behavior, which lead to wrong commands to the downstream cobot control pipeline, and even result in dangerous actions. There are several research directions that we can push forward in order to address this safety concern of more advanced collaborative robot systems.

First, we need to develop more accurate and faster human tracking algorithms and human behavior models. Inference speed is essential to improve the latency of the cobot execution. Increase of frequency helps mitigate the influence of outliers and capture sudden changes to make the cobot highly reactive to safety-critical situations. Second, we need to develop programs to help the cobot clarify ambiguity in human intentions. Different human intentions may induce similar human actions, and the cobot should distinguish between these intentions either implicitly by acting in a way that expects the human to react differently for different intentions, or explicitly by querying the human by speech. Third, we need to develop vision and force/torque fusion algorithms for emergency detection. Abnormal contacts should be classified into emergent and less critical ones based on checking by vision whether the contact is between the cobot and the human or between the cobot and objects, so the cobot can act less conservatively with safety guaranteed. Fourth, we need to develop a logging pipeline for abnormality/accident investigation and dataset collection with the purpose of customizing cobot skills. In certain industrial tasks, especially free-form ones, we argue that cobot with skills customized for its human partner could lead to safer and more efficient collaboration.

\bibliographystyle{IEEEtran}
\bibliography{bib}

\end{document}